\documentclass{article} % For LaTeX2e

\usepackage{times}
\usepackage{graphicx} % more modern
\usepackage{subfigure}

\usepackage{natbib}

\usepackage{algorithm}
\usepackage{algorithmic}

\usepackage{hyperref}

\usepackage[accepted]{icml2016}

\usepackage{natbib}
\usepackage{amsmath}
\usepackage{amsthm}
\usepackage{amssymb}
\usepackage{relsize}
\usepackage{xfrac}

\graphicspath{{.}}

\renewcommand{\vec}[1]{{\mathbf{#1}}}

\newcommand{\E}[2]{\mathop{\mathlarger{\mathbb{E}} }_{#1}\left[#2\right]}

\newcommand{\qrob}[2]{q(#1 \, | \, #2)}

\newcommand{\BB}{{\cal B}}
\newcommand{\DD}{{\cal D}}

\newcommand{\kk}{{(k)}}

\newcommand{\ps}{p^*}
\newcommand{\pts}{\tilde{p}^*}
\newcommand{\x}{\vec{x}}

\newcommand{\h}{\vec{h}}

\newcommand{\vect}[1]{\boldsymbol{\mathbf{#1}}}

\usepackage[normalem]{ulem}
\usepackage{color}

\title{Bidirectional Helmholtz Machines}

\author{
J\"org Bornschein,
Samira Shabanian, Asja Fischer, Yoshua Bengio$^{*}$ \\
Dept. Computer Science and Operations Research, University of Montreal \\
$^{*}$ Yoshua Bengio is a CIFAR Senior Fellow \\
\texttt{\{bornj,shabanis,fischer,<findme>\}@iro.umontreal.ca}
}

\begin{document}

\maketitle
\begin{abstract}

Efficient unsupervised training and inference in deep generative models remains
a challenging problem. One basic approach, called Helmholtz machine or Variational Autoencoder, involves
training a top-down directed generative model together with a bottom-up
auxiliary model used for approximate inference.
Recent results indicate that better generative models can be obtained with better approximate inference procedures.
Instead of improving the inference procedure, we here propose a new model, the bidirectional Helmholtz machine,
which guarantees that the top-down and bottom-up distributions can efficiently
invert each other.
We achieve this by interpreting both the top-down and the bottom-up directed models as approximate inference distributions and by defining the model
distribution to be the geometric mean of these two.
We present a lower-bound for the likelihood of this model and we show that optimizing this bound
regularizes the model so that the Bhattacharyya distance between the bottom-up and top-down
approximate distributions is minimized.
This approach results in state of the art generative models which prefer significantly deeper
architectures while it allows for orders of magnitude more efficient likelihood estimation.   % XXX
\end{abstract}

\section{Introduction and background}
\label{sec:introduction}

Training good generative models and fitting them to complex and high
dimensional training data with probability mass in multiple disjunct locations
remains a major challenge. This is especially true for models with multiple
layers of deterministic or stochastic variables, which is unfortunate because it
has been argued previously~\citep{Hinton06,Bengio-2009-book} that deeper
generative models have the potential to capture higher-level abstractions and
thus generalize better. Although there has been progress in dealing with
continous-valued latent variables~\citep{Kingma+Welling-ICLR2014}, building a hierarchy of representations,
especially with discrete-valued latent variables, remains a challenge.

With the Helmholtz machine \citep{Hinton95,Dayan-et-al-1995},
a concept was introduced that proposed to not only fit a powerful but
intractable generative model $p(\x, \h)$ to the training data, but also to
jointly train a parametric approximate inference model $q(\h|\x)$. The distribution $q$ is
used to perform approximate inference over the latent variables $\h$ of
the generative model given an observed example $\x$, i.e. to approximate $p(\h|\x)$. This basic idea has been
applied and enhanced many times; initially with the wake-sleep algorithm
(WS, \citet{Hinton95,Dayan1996varieties}) and more recently with the variational autoencoder (VAE,
\citet{Kingma+Welling-ICLR2014}), stochastic backpropagation and approximate
inference in deep generative models~\citep{Rezende-et-al-ICML2014}, neural
variational inference and learning\ (NVIL, \citet{Andriy-Karol-2014}) and
reweighted wake-sleep (RWS, \citet{BornscheinBengio2015}).
%
% \jbrm{Most of these approaches rely on the variational bound to perform
% appropriate inference and to obtain an objective function that contains the
% parameters of both the generative model $p$ and the approximate inference model
% $q$ in one joint expression (e.g. WS, VAE and NVIL).}
%

Recent results indicate that significant improvements can be made when better
approximate inference methods are used: \citet{Salimans-et-al-arxiv2014} for example %Saliman et. al. (2014) for example
presented an iterative inference procedure that improves the samples from $q$ by
employing a learned MCMC transition operator. In \citep{hjelm2015iterative} the authors propose an
iterative, gradient based procedure to refine the approximate posterior. \citet{burda2015importance}
present the importance weighted auto encoder (IWAE), an improved VAE that, similarly to
RWS, uses multiple samples from $q$ to calculate gradients. And RWS already reported that
autoregressive $q$ distributions lead to noticable improvements.
%
%
% In \citep{burda2015importance}, an improved
% VAE is presented that incorporates ideas from RWS. % add Russ's reweighted thingy, probably
% Not directly related to the Helmholtz machine but similar in spirit is the
% approach described in \citep{DicksteinEtAl2015}.
%It also relies on a bottom-up distribution $q(\h|\x)$ to perform approximate inference for the latent
%variables in a very deep generative model $p(\x, \h)$. But the inference
%distribution $q$ is fixed and only the parameters of $p$ are learned.
%
%
In contrast to these previous approaches that aimed at incorporating more
powerful inference methods to gain better approximations of $p(\h|\x)$, we here propose to regularize the top-down model $p$
such that its posterior stays close to the approximate inference distribution
$q$ (and vice versa). We archive this by interpreting both $p$ and $q$ as approximate inference
models for our actual generative model $\ps$, which is defined to be the
geometric mean over the top-down and bottom-up approximate inference models,
i.e., $\ps(\x, \h) = \sfrac{1}{Z} \sqrt{p(\x, \h) q(\x, \h)}$.

%
%In Section \ref{sec:theconstruction} we will introduce the model in detail and
%discuss important theoretical properties. We will show that the proposed
In Section \ref{sec:theconstruction} we show that this model definition leads to %\af{introduce the model in more detail and show that maximizing the log-likelihood of the model} leads to
an objective that can be interpreted as using a regularization term that
encurages solutions where $p$ and $q$ are close to each other in terms of the Bhattacharyya distance. In Section \ref{sec:inference} we will explain how to
perform importance sampling based training and inference. The ability to model
complex distributions and the computational efficiency of this approach are
demonstrated empirically in Section \ref{sec:experiments}.  %
% The motivation behind this definition is to ensure that the intractable
% generative model $\ps$ stays close to the approximate inference models we have
% at our disposal. In fact, we show that the proposed objective can be interpreted
% as adding a regularization term to the log-likelihood objective towards solutions
% where $p$ and $q$ are close to each other in terms of the Bhattacharyya distance.

\section{Model definition and properties}
\label{sec:theconstruction}

We introduce the bidirectional Helmholtz Machine (BiHM) by defining a joint probability distribution over
three variable vectors, an observed vector $\x$ and two latent variable
vectors $\h_1$ and $\h_2$.
Analogous to a Deep Boltzmann Machine (DBM,~\citet{Salakhutdinov2009deep}),
we think of these as layers in a neural network with links between $\x$ and
$\h_1$ on the one side, and $\h_1$ and $\h_2$ on the other side.
%\af{NOTE: Do we want to change this statement, because we found a different structure than in DBM?.}
We will present our approach for the specific case of an architecture with two hidden layers,
but it can be applied to arbitrary graphs of variables
without loops. It can especially be used to train architectures with more than two
stacked layers of latent variables.

Let $\ps(\x, \h_1, \h_2)$ be a joint probability distribution constructed in a
specific way from two constituent distributions $p(\x, \h_1, \h_2)$ and
$q(\x, \h_1, \h_2)$,
\begin{align*}
  p^*(\x, \h_1, \h_2) &= \frac{1}{Z} \sqrt{p(\x, \h_1, \h_2) \; q(\x, \h_1, \h_2)}\nonumber \enspace,
 % \label{eqn:ps-def}
\end{align*}
where $Z$ is a normalization constant and $p$ and $q$ are directed graphical models from $\h_2$ to
$\x$ and vice versa,
\begin{align*}
  p(\x, \h_1, \h_2) &= p(\h_2) \; p(\h_1 | \h_2) \; p(\x  | \h_1) \text{\;\;\; and \;\;\;} \nonumber \\
  q(\x, \h_1, \h_2) &= q(\x)   \; q(\h_1 | \x)   \; q(\h_2 | \h_1) \nonumber \enspace.
 % \label{eqn:direct-def}
\end{align*}
We assume that the prior distribution $p(\h_2)$ and all conditional
distributions belong to parametrized families of distributions which can
be evaluated and sampled from efficiently. For $q(\x)$ we do not assume an
explicit form but define it to be the marginal
\begin{align*}
  q(\x) &= \ps(\x) = \sum_{\h_1, \h_2} \ps(\x, \h_1, \h_2) \\
        &= \frac{\sqrt{q(\x)}}{Z}
          \sum_{\h_1, \h_2} \sqrt{p(\x, \h_1, \h_2) \; q(\h_1 | \x) q(\h_2 | \h_1)}
         \\
        &= \Big( \frac{1}{Z}
            \sum_{\h_1, \h_2} \sqrt{p(\x, \h_1, \h_2) \; q(\h_1 | \x) q(\h_2 | \h_1)}
           \Big)^2.
\end{align*}
The normalization constant $Z$ guarantees
that $\sum_{\x,\h_1,\h_2} \ps(\x, \h_1, \h_2)=1$.
Using the Cauchy-Schwarz inequality
%$\left| \int f(y)g(y) dy \; \right|^2 \le \int \left| f(y) \right|^2
%dy \int \left| g(y') \right|^2 dy'$
$| \sum_y f(y)g(y) |^2 \le \sum_y \left| f(y) \right|^2 \times
\sum_y  \left| g(y) \right|^2 $
and identifying $\sqrt{p(\x, \h_1, \h_2)}$ with
$f(y)$ and $\sqrt{q(\x, \h_1, \h_2)}$ with $g(y)$, it becomes clear that
$Z = \sum_{\x, \h_1, \h_2} \sqrt{p(\x, \h_1, \h_2)q(\x, \h_1, \h_2)}  \le 1$
for arbitrary $p$ and $q$. Furthermore, we see that $Z=1$ if and only if $p(\x, \h_1, \h_2) = q(\x, \h_1, \h_2)$.
We can therefore obtain a lower bound on the marginal probability $\ps(\x)$
by defining
\begin{align}
  \pts(\x) &= \Big(\sum_{\h_1, \h_2} \sqrt{p(\x, \h_1, \h_2) q(\h_1 | \x) \; q(\h_2 | \h_1)} \Big)^2    \nonumber
   \\  &= Z^2 \; \ps(\x) \le \ps(\x) \enspace.
   \label{eqn:pts-def}
\end{align}
This suggests that the model distribution $\ps(\x)$ can be fitted to some
training data by maximizing the bound of the log-likelihood (LL) $\log \pts(\x)$ instead
of $\log \ps(\x)$, as we elaborate in the following section.  Since $\log \pts(\x)$
can reach the maximum only when $Z \to 1$, the model is implicitly pressured to
find a maximum likelihood solution that yields $p(\x, \h_1, \h_2) \approx q(\x,
\h_1, \h_2) \approx \ps(\x, \h_1, \h_2)$.

\subsection{Alternative view based on the Bhattacharyya distance}

%\afrm{Let us consider the case where we would actually maximize $\log \ps(\x)$
%instead of the lower bound $\log \pts(\x)$. We can then decompose the objective}
Recalling the Bhattacharyya distance $D_B(p, q) = -\log \sum_y
\sqrt{p(y)q(y)}$ (for which holds $D_B(p, q) \ge 0$ for arbitrary distributions $
p, q $ and $D_B(p, q) = 0$ only if $p= q$) the model LL $\log
\ps(\x)$ can be decomposed into % two terms
\begin{align}\
   \log \ps(\x)
    &= 2 \log \sum_{\h_1, \h_2} \sqrt{p(\x, \h_1, \h_2) q(\h_1|\x) q(\h_2|\h_1)}
        \nonumber \\
    & \hspace{0.5cm} - 2 \log \sum_{\x', \h'_1, \h'_2} \sqrt{p(\x', \h'_1, \h'_2) q(\x', \h'_1, \h'_2)} \
        \nonumber \\
    %&= \log \pts(\x) - 2 \log Z \nonumber \\
    &= \log \pts(\x) + 2 \, D_B(p, q)
     \ge \log \pts(\x) \enspace, \label{eqn:bound} % \\
    %\text{with }  D_B(p, q&) \ge 0 \text{ for arbitrary } p, q  %\nonumber \\
    %\text{\; and \;} D_B(p, q) = 0 \text{ only when } p = q %\text{.} \nonumber
\end{align}
where we clearly see that the proposed training objective $\log \pts(\x)$ corresponds to the % correct (but intractable)
LL $\log \ps(\x)$ minus 2 times the Bhattacharyya distance $D_B(p, q)$,
i.e., it is maximzing the true LL and minimizing the distance between $p$ and $q$.
We can compare this to the variational approach, where the marginal probability
$\log p(\x)$ of some model containing latent variables $\h$ is rewritten in terms
of the KL-divergence $D_{KL}(q(\h|\x)\,||\, p(\h|\x))=\sum_h q(\h|\x) \log \frac{q(\h|\x)}{p(\h|\x)}
\ge 0$ to obtain a lower bound
\begin{align}
  %\log p(y) = \underbrace{\E{z \sim q(z|y)}{\log p(y, z) - \log q(z|y)}}_{lower bound}
  \log p(\x) &= \E{\h \sim q(\h|\x)}{\log p(\x, \h) - \log q(\h|\x)} \nonumber \\
           & \hspace{1.0cm} + D_{KL}(q(\h|\x)\,||\,p(\h|\x)) \nonumber \\
           &\ge \E{\h \sim q(\h|\x)}{\log p(\x, \h) - \log q(\h|\x)}
            \label{eqn:klbound} \enspace .
\end{align}
Analogous to variational methods that maximize the lower bound \eqref{eqn:klbound}, we can thus
maximize $\log \pts(\x)$, and it will tighten the bound as $D_B(p, q)$
approaches zero. While this seems very similar to the variational lower bound, we
should highlight that there are some important conceptual differences: 1) The
KL-divergence in variational methods measures the distance between
distributions {\em given some training data}. The Bhattacharyya distance here
in contrast quantifies a property of the model $\ps(\x, \h_1, \h_2)$
independently of any training data.  In fact, we saw that $D_B(p, q) = -\log
Z$.  2) The variational lower bound is typically used to construct approximate
inference algorithms. We here use our bound $\pts(\x)$ just to remove the
normalization constant $Z$ from our target distribution $\ps(\x)$.
Even after applying the lower-bound, we still have to tackle the inference
problem which manifests itself in form of the full combinatorial sum over
$\h_1$ and $\h_2$ in equation \eqref{eqn:pts-def}. Although it seems intuitively
reasonable to use a variational approximation on top of the bound $\pts(\x)$
%\af{[Check: Do we still think this after what we found on the bord (it may get equivalent to VAE)?]}, we
we will here not follow this direction but rather use importance sampling to
perform approximate inference and learning (see section \ref{sec:inference}).
Combining a variational method with the bound $\pts(\x)$ is therefore subject to
future work.

We can also argue that optimizing $\log \pts(\x)$ instead of $\log \ps(\x)$ is
beneficial in the light of the original goal we formulated in section
\ref{sec:introduction}: To learn a generative model $\ps$ that is
regularized to be close to the model $q$ which we use to perform
approximate inference for $\ps$. Let us assume we have two equally well
trained models $\ps_{\vect{\theta}_1}$ and $\ps_{\vect{\theta}_2}$, i.e.,
in expectation over the empirical distribution $\E{}{\log
\ps_{\vect{\theta}_1}(\x)} = \E{}{\log \ps_{\vect{\theta}_2}(\x)}$, but the expected bound
$\pts(\x)$ for the first model is closer to the LL than
the expected bound for the second model: $\E{}{\log \pts_{\vect{\theta}_1}(\x)} > \E{}{\log
\pts_{\vect{\theta}2}(\x)}$. Using equation \eqref{eqn:bound} we see that
$D_B(p_{\vect{\theta}_1}, q_{\vect{\theta}_1}) < D_B(p_{\vect{\theta}_2}, q_{\vect{\theta}_2})$ which
indicates that $q_{\vect{\theta}_1}$ is closer to $\ps_{\vect{\theta}_1}$ than $q_{\vect{\theta}_2}$
is to $\ps_{\vect{\theta}_2}$ (when we measure their distance using the Bhattacharyya
distance). According to our original goal, we thus prefer solution
$\ps_{\vect{\theta}_1}$, where the bound $\pts(\x)$ is maximized and the distance
$D_B(p, q)$ minimized.

Note that the decomposition \eqref{eqn:bound} also emphasizes why our recursive
definition $q(\x) = \sum_{\h_1, \h_2} \ps(\x, \h_1, \h_2)$ is a consistent and
reasonable one: minimizing $D_B(p, q)$ during learning means that the joint
distributions $p(\x, \h_1, \h_2)$ and $q(\x, \h_1, \h_2)$ approach each
other. This implies that the marginals $p(\h_l)$ and $q(\h_l)$ for all layers
$l$ become more similar. This also implies $p(\x) \approx q(\x)$ in the limit of
$D_B(p, q) \to 0$; a requirement that most {\em simple} parametrized
distributions $q(\x)$ could never fulfill.

% \af{[Check: Could we leave the following sentence out?]}
% It is thus a
% piece of good fortune that the optimal, but intuitively clumsy definition
% $q(\x) = \sum_{\h_1, \h_2} \ps(\x, \h_1, \h_2)$ results in the likelihood
% expression \afrm{\eqref{eqn:pts-def}} \af{\ref{eqn:bound}}, an expression that has no higher computational
% complexity and that has mostly the same basic structure as the likelihood expression
% for any other model containing latent variables.

\begin{figure*}[t]
 \includegraphics[width=\linewidth]{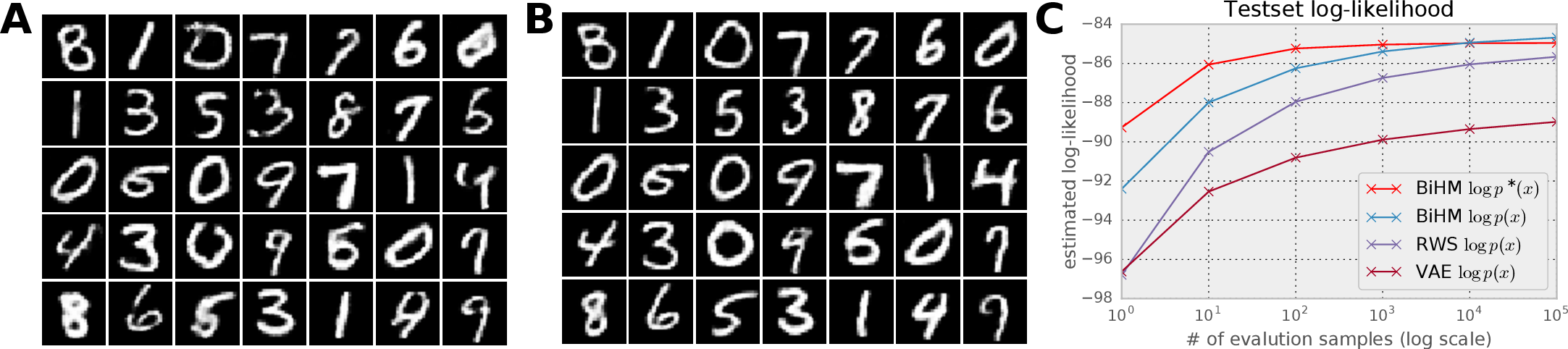} \\
 \vspace{-0.8cm}
 \caption{MNIST experiments: {\bf A}) Random samples from top-down model $p(\x)$.
   {\bf B})  Generally improved samples after running $10$ iterations of Gibbs sampling to obtain approximate samples from the joint model $\ps(\x)$.
  %In {\bf A} and {\bf B} we show expected samples instead of sampling from the bottom-most Bernoulli distribution, as usual.
  In {\bf A} and {\bf B} we show expected samples instead of sampling from the bottom Bernoulli distribution, as usual.
  %(gray level = Bernoulli probability).
  %\af{[Check: The expected samples are gained by trash-holding?]}
  {\bf C}) Sensitivity of the test set LL estimates to the number of samples $K$.
  We plot the test set $\log p(\x)$ and $\log \ps(\x)$ estimates of our best BiHM model together with
  the $\log p(\x)$ estimates of our best RWS and VAE models.
  % Histogram of importance weights when using $q(\h|\x)$ as proposal for $\ps$
  % (when performing inference for a given $\x$ from the test set) and when using
  % $p$ as proposal for $\ps$ (when drawing samples according to algorithm
  % \af{[Check: This could form a small IS analysis section, together with the results for the estimated sampling size.]}
  %\ref{alg:resampling}).
  }
 \label{fig:mnist}
\end{figure*}

\begin{figure*}[t]
 \begin{centering}
   \includegraphics[width=0.95\linewidth]{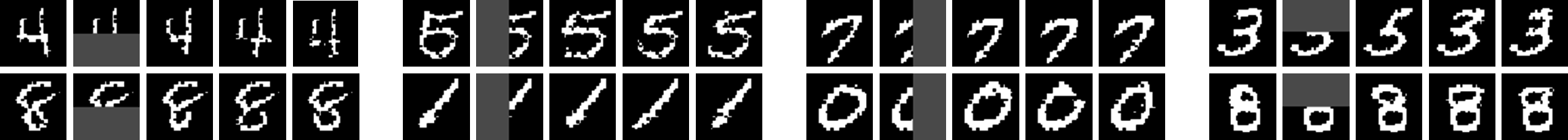} \\
 \end{centering}
 \vspace{-0.3cm}
 \caption{Inpainting of binarized MNIST digits. The left column in each block shows the original digit
  randomly sampled from the MNIST test set; the first column shows the masked version presented to the algorithm,
  and the next three columns show different, independent reconstructions after 100 Gibbs iterations.
  (see section \ref{sec:inpainting}). All images were selected randomly.
  }
 \label{fig:inpainting}
\end{figure*}

\section{Inference and training} % with importance sampling}
\label{sec:inference}

\begin{algorithm}[t]
\caption{Training $\ps(\x)$ using $K$ importance samples}  %with $q$ as proposal }
\label{alg:train}
\begin{algorithmic}
\FOR{number of training iterations}
  \STATE{$\bullet$ Sample
 % example(s)
  $\x$  from the training distribution (i.e.\,$\x \sim \DD$)}
  \FOR{$k= 1, 2, \dots, K$}
    \STATE{$\bullet$ Sample $\h^{(k)}_1 \sim q(\h^{(k)}_1|\x ); \h^{(k)}_l \sim q(\h^{(k)}_l|\h^{(k)}_{l-1})$ \\
      \hspace{0.5cm} (for layers $l=2$ to $L$) }
    %   for each layer $l=2$ to $L$ sample $ \h^{(k)}_l \sim q(\h^{(k)}_l|\h^{(k)}_{l-1})$}
    % \STATE{$\bullet$
    % Layerwise sample latent variables:
     %(first layer above $\xVec$, second layer, etc. up to top hidden layer).}
    \STATE{$\bullet$ Compute $q(\h^{(k)}|\x)$ and $p(\x,\h^{(k)})$ \\
      \hspace{0.5cm} (for $\h^{(k)}=(\h^{(k)}_1,\dots, \h^{(k)}_L)$)}
  \ENDFOR
  \STATE{$\bullet$ Compute unnormalized importance weights \\
    \hspace{0.5cm} $\omega_k = \sqrt{\sfrac{p(\x, \h^{(k)})}{q(\h^{(k)}|\x)}}$}
  \STATE{$\bullet$ Normalize the weights $\tilde{\omega}_k = \sfrac{\omega_k}{\sum_{k'} \omega_{k'}}$}
  %\STATE{$\bullet$ Compute unbiased likelihood estimator $\hat{p}(\x) = {\rm average}_k \; \omega_k$}
  %\STATE{$\bullet$ Compute log-likelihood estimator $\widehat{\LL}(\x) = \log {\rm average}_k \; \omega_k$}
  \STATE{$\bullet$ Update parameters of $p$ and $q$: gradient descent \\
    %\hspace{0.5cm} gradient estimator $2 \sum_k \tilde{\omega}_k \frac{\partial \log \ps(\x,\h^{(k)})}{\partial \theta}$}
    \hspace{0.3cm} with gradient estimator \\
    \hspace{0.5cm} $\sum_k \tilde{\omega}_k \frac{\partial}{\partial \theta}  \log p(\x, \h^\kk) q(\h^\kk|\x)$}

\ENDFOR
%\af{[Check: Can we describe the last step more accurate? Do we need to care about that this would be a online and not the mini-batch version?]}
\end{algorithmic}
\end{algorithm}

Based on the construction of $\ps(\x)$ outlined in the previous section we can
define a wide range of possible models. Furthermore, we have a wide range of
potential training and appropriate inference methods we could employ to
maximize $\log \pts(\x)$.

In this text we concentrate on binary latent and observed variables $\x,\h_1, \h_2$ %\in \{0, 1\}^D$
 and model all our conditional distributions by simple
sigmoid belief network layers, e.g.,
$ p(\x\,|\,\h_1) = \prod_i \BB(x_i \, | \, \sigma(W_i \; \h_{1} + b_i) ) $
where $\BB(x_i\,|\,c)$ refers to the Bernoulli distribution with \mbox{$P(x_i=1)=c$,
$W_i$} are the connection weights between the latent variables $\h_1$ and the
visible variable $x_i$; $b_i$ is the bias of $x_i$, and $\sigma(\cdot)$ is the
sigmoid function. For our top-level prior $p(\h_2)$, we use a factorized
Bernoulli distribution: $p(\h_2) = \prod_i \BB(h_{2,i}\,|\,\sigma(b_{2,i}))$.

% \af{[Check: We could change the order here. First introducing how we can train
% the model/approximate the gradient, and then telling how we can evaluate the
% likelihood/model. In this context we could also explain how to estimate Z. And
% by the way, do we estimate $log \ps$ and $\log p$ based on  (7) and (8)? And
% don't we get a lower bound by this again?]}

We form an estimate  of $\pts(\x)$ by using importance sampling instead of the
exhaustive sum over $\h_1$ and $\h_2$ in equation \eqref{eqn:pts-def}.
We use $q(\h_1|\x) q(\h_2|\h_1)$ as the proposal distribution which is by
construction easy to evaluate and to sample from:
\begin{align}
  \pts(\x) &= \Big( \sum_{\h_1, \h_2} \sqrt{p(\x, \h_1, \h_2) \; q(\h_1 | \x) \; q(\h_2 | \h_1)} \Big)^2 \nonumber \\
           &= \Big( \E{ \substack{\h_2 \sim q(\h_2|\h_1) \\ \h_1 \sim q(\h_1|\x)}}
                 {\sqrt{ \frac{p(\x, \h_1, \h_2)}{q(\h_1 | \x) \; q(\h_2 | \h_1)}}}
            \Big)^2 \nonumber \\
           &\simeq \Big(
              \frac{1}{K} \sum_{k=1}^K \sqrt{ \frac{p(\x, \h_1^\kk, \h_2^\kk)}{q(\h_1^\kk | \x) \; q(\h_2^\kk | \h_1^\kk)}}
            \Big)^2 \label{eqn:phs}
\end{align}
          %  &\text{ \; with }
          %   \begin{array}{c}
          %          \h_1^\kk \sim q(\h_1|\x)\;\; \\
          %          \h_2^\kk \sim q(\h_2|\h_1^\kk)
          %   \end{array} \text{\;.} \nonumber
%
with $ \h_1^\kk \sim q(\h_1|\x)$ and $\h_2^\kk \sim q(\h_2|\h_1^\kk)$.
Using the same approach, we can also derive the well known estimator for
the marginal probability of a datapoint under the top-down generative model $p$:
\begin{align}
  p(\x) &= \E{ \substack{\h_2 \sim q(\h_2|\h_1) \\ \h_1 \sim q(\h_1|\x)}}
                 {\frac{p(\x, \h_1, \h_2)}{q(\h_1 | \x) \; q(\h_2 | \h_1)}}  \nonumber \\
    &\simeq \frac{1}{K}
        \sum_{k=1}^K  \frac{p(\x, \h_1^\kk, \h_2^\kk)}{q(\h_1^\kk | \x) \; q(\h_2^\kk | \h_1^\kk)}
        \enspace.\label{eqn:ph}
\end{align}
Comparing \eqref{eqn:phs} and \eqref{eqn:ph} and making use of Jensen's inequality it becomes clear that $p(\x)\geq \pts(\x)$.

Analogous to the parameter updates in RWS
\citep{BornscheinBengio2015}, we can derive an importance sampling based estimate
for the %parameter gradients
LL gradient with respect to the parameters of $p$ and $q$
(jointly denoted by $\vect{\theta}$) and use it to optimize our objective
(we use $\h$ to jointly denote the latent variables of all layers) % \af{in the following derivation}).
% (see Appendix \ref{sec:AppendixGradient} for more details):
%
%\begin{align}
%   \frac{\partial}{\partial \vect{\theta}} \log & \pts(\x) \simeq \label{eqn:grad-est} \\
%    & 2 \sum_{k=1}^K \tilde{\omega}_k
%            \frac{\partial}{\partial \vect{\theta}} \log \sqrt{p(\x, \h_1^{(k)}, \h_2^{(k)}) q(\h_1^{(k)}, \h_2^{(k)}|\x)}
%    \nonumber
%\end{align}
%with samples $\h_1^\kk \sim \qrob{\h_1}{\x}  , \h_2^\kk \sim \qrob{\h_2}{\h_1^\kk} $, for $k=1,\dots, K$,
% and  importance weights
%\begin{equation}
% \tilde{\omega}_k = \frac{\omega_k}{\sum_{k'} \omega_{k'}} , \text{where} \, \enspace
%         \omega_k = \sqrt{ \frac{p(\x, \h_1^\kk,\h_2^\kk)}{q(\h_1^\kk, \h_2^\kk|\x)} } \;
%               \text{\;.}
%\end{equation}
%
\begin{align}
  \frac{\partial}{\partial \vect{\theta}} &\log \pts(\x)
    = \frac{\partial}{\partial \vect{\theta} }\log \Big( \sum_\h \sqrt{p(\x, \h) q(\h|\x) } \Big)^2 \nonumber \\
%  &=2 \left(\sum_{\h}  \sqrt{p(\x, \h) q(\h|\x) } \right)^{-1}
    &= \frac{
     2 \sum_{\h} \sqrt{p(\x, \h) q(\h|\x)}
       \frac{\partial}{\partial \vect{\theta}} \log \sqrt{p(\x, \h) q(\h|\x)}
    }{\sum_{\h'}  \sqrt{p(\x, \h') q(\h'|\x) }} \nonumber \\
       &\simeq \sum_{k=1}^K \tilde{\omega}_k
           \frac{\partial}{\partial \vect{\theta}} \log p(\x, \h^{(k)}) q(\h^{(k)}|\x)
           \label{eqn:grad-est} \enspace,
   \end{align}
  with $\h^\kk \sim \qrob{\h}{\x}$ and importance weights
\begin{equation}
 \tilde{\omega}_k = \frac{\omega_k}{\sum_{k'} \omega_{k'}} \, \text{ where } \, \enspace
         \omega_k = \sqrt{ \frac{p(\x, \h^\kk)}{q(\h^\kk|\x)} } \;
               \text{\;,} \nonumber
\end{equation}
 for $k=1,\dots, K$.
%\text{with samples \; } & \h^\kk \sim \qrob{\h}{\x} \; \text{for\ } k=1,\dots, K
%\text{\; and importance weights  }\nonumber \\
%%\text{importance weights \; }
%              \tilde{\omega}_k &= \frac{\omega_k}{\sum_{k'} \omega_{k'}} \text{\;, with \;\;} \omega_k = \sqrt{ \frac{p(\x, \h^{(k)})}{\qrob{\h^{(k)}}{\x}} }  \enspace. \nonumber
%

In contrast to VAEs and IWAEs, the updates do not require any form of backpropagation through layers
because, as far as the gradient computation $\frac{\partial}{\partial \vect{\theta}}
\log \sqrt{p(\x, \h^\kk) q(\h^\kk|\x)}$ is concerned, these samples are considered
fully observed.
%\af{[Check: I am not sure if that is what the following sentence aimed to say. And what does normalized mean? And do we need to say this anyway, because it is the standard procedure to average over a batch? ]}
The gradient approximation \eqref{eqn:grad-est} computes the weighted average
over the individual gradients. These properties are basically inherited from the RWS training
algorithm.
But in contrast to RWS, and in contrast to most other algorithms which employ
a generative model $p$ and an approximate inference model $q$, we here automatically
obtain parameter updates for both $p$ and $q$ because we optimize $\ps$ which contains
both. The resulting training method is summarized in algorithm \ref{alg:train}.

\paragraph{Estimating the partition function $Z$}
To compute $\ps(\x) = \frac{1}{Z^2} \pts(\x)$ and to monitor the training progress
it is desirable to estimate the normalization constant $Z$. In stark contrast to undirected
models like RBMs or DBMs, we can here derive an {\em unbiased} importance sampling based estimator for $Z^2$:
\begin{align}
  Z^2 &= \E{\x, \h \sim p(\x, \h)}
     {\sqrt{\frac{q(\h|\x)}{p(\x, \h)}}
         \E{\h' \sim q(\h|\x)}
           {\sqrt{\frac{p(\h', \x)}{q(\h'|\x)}}}
     } \nonumber \\
  &= \E{\substack{\x, \h \sim p(\x, \h) \\ \h' \sim q(\h'|\x)}}
      {\sqrt{\frac{p(\x, \h') q(\h|\x) }{p(\x, \h) q(\h'|\x)}}} \enspace.
  \label{eqn:z-est}
\end{align}
%\af{where $h$ is used to summarize the state of the variables in all hidden layers.}
%
We denote the number of samples used to approximate the outer expectation and the
inner expectation with $K_\text{outer}$ and $K_\text{inner}$ respectively.
%number of samples \af{used for approximating}  the inner expectation $K_\text{inner}$.
%and the total number of samples with $K=K_{outer}*K_{inner}$.
In the experimental section, we show that we obtain
%the computationally most efficient estimator with $K_\text{inner} = 1$,
%\af{[Check: Doesnt this sound strange at this point? Of course taking 1 sample compared to taking many is efficient....]}
high quality estimates for $Z^2$ with $K_{\text{inner}}$=1 and a relatively small number of samples $K_{\text{outer}}$.
By taking the logarithm, we obtain a biased estimator for $2 \log Z$, which will,
unfortunately, underestimate $2 \log Z$ on average due to the concavity of
the logarithm and the variance of the $Z^2$ estimate.
% \af{[Check: Is this really due to the variance?]} %YB: yes, with zero variance the bias would also xbe 0!
This can lead to
overestimates for $\log \ps(\x)$ (see equation \eqref{eqn:phs}) if we are not
careful. Fortunately, the bias on the estimated $\log Z$ is induced only by the concavity of the logarithm;
the underlying estimator for $Z^2$ is unbiased.
% \af{[Check: The variance does not effect the bias! We have to work over the following part!]}
% It does! We are talking about the variance of Z^2 influencing the biais of the estimate of log Z.
We can thus effectively minimize the bias by minimizing the
variance of the $Z^2$ estimate (e.g. by taking more samples). This is a much
better situation than for $Z$-estimating methods that rely on Markov
chains in high dimensional spaces, which might miss entire modes because of
mixing issues.

\begin{figure*}[t]
 %\vspace{-0.4cm}
 %\begin{centering}
     \includegraphics[width=1.0\linewidth]{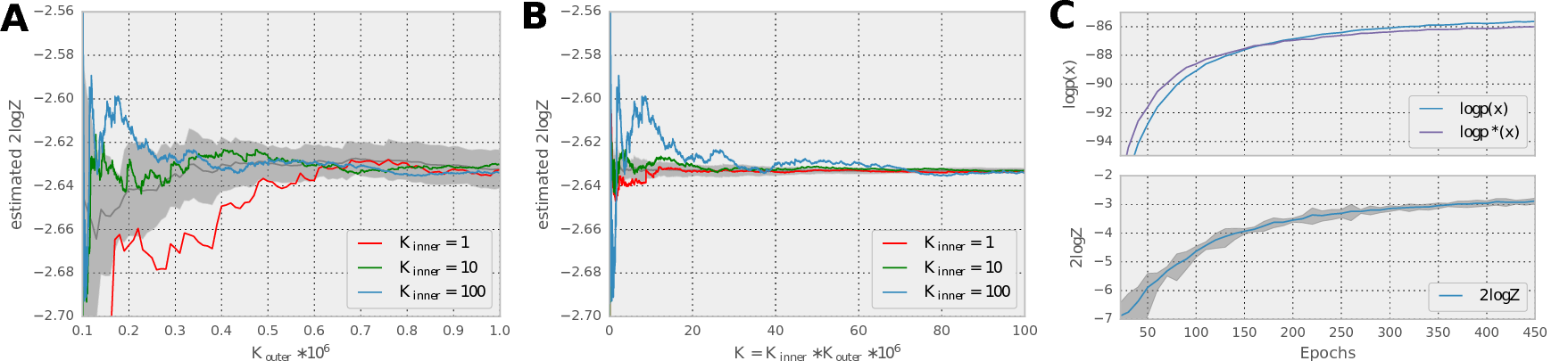} \\
 %\end{centering}
 \vspace{-0.7cm}
 \caption{$\log Z^2$ estimates for different values of $K_{\text{inner}}$ as a function of
 {\bf A}) the number of samples $K_{\text{outer}}$,
 {\bf B}) the total number of samples $K_{\text{inner}} \cdot K_{\text{outer}}$ for the BiHM trained on MNIST;
 the gray region shows the mean and the standard deviation for 10 runs with $K_\text{inner}$=1.
 This shows that, from the point of view of total computation, convergence is fastest with $K_\text{inner}$=1; and that
 we obtain a high quality estimate of the partition function with only a few million samples.
 {\bf C}) Evolution of the estimates of $\log p(\x)$, $\log \ps(\x)$, and
 $2 \log Z$ during training on MNIST.}
 \label{fig:zest}
\end{figure*}

\begin{table*}[t]
\label{table:uci}
\centering
%\scriptsize
%\tiny
\footnotesize
\begin{tabular}{|l|c|c|c|c|c|c|c|c|}
\hline
Model         & \tiny \bf ADULT & \tiny \bf CONNECT4 & \tiny \bf DNA & \tiny \bf MUSHROOMS & \tiny \bf NIPS-0-12 & \tiny \bf OCR-LETTERS & \tiny \bf RCV1 & \tiny \bf WEB \\
\hline
\hline
\footnotesize
{\bf auto regressive}   &          &           &             &             &        &       &       & \\
%\;{FVSBN}                & 13.17    & 12.39     & 83.64       & 10.27       & 276.88 & 39.30 & 49.84 & 29.35 \\
\;NADE                  & 13.19    & 11.99     & 84.81       &  9.81       & 273.08 & 27.22 & 46.66 & 28.39 \\
\;EoNADE                & 13.19    & 12.58     & 82.31       &  9.68       & 272.38 & 27.31 & 46.12 & 27.87 \\
\;DARN                  & 13.19    & 11.91     & 81.04       &  9.55       & 274.68 & 28.17 & 46.10 & 28.83 \\
\;RWS - NADE            & 13.16    & 11.68     & 84.26       &  9.71       & 271.11 & 26.43 & 46.09 & 27.92 \\
\hline
{\bf non AR}                            &          &           &             &             &        &       &       & \\
\;{RBM}                                 & 16.26    & 22.66  & 96.74      & 15.15       & 277.37    & 43.05      & 48.88     & 29.38 \\
\;RWS - SBN                             & 13.65    & 12.68  & 90.63      &  9.90       & 272.54    & 29.99      & 46.16     & 28.18 \\
%\;{\tiny hidden units}       & 5-20-100 & 10-50-150 & 10-150      & 10-50-150   & 10-50-150 & 10-100-300 & 10-50-200 & 10-50-300 \\
\hline
{\bf BiHM}                &       &         &         &        &        &       &       & \\
%\;$-\log \tilde{p}(\x)$       & 13.58 &  11.98  &  86.33  &  9.34  & 270.33 & 27.91 & 45.67  & 28.02 \\
%\;$-\log \tilde{\ps}(\x)$     & 13.79 &  12.12  &  86.39 &  9.4   & 272.62 & 29.46 & 45.78 & 28.78 \\
\;-$\log p(\x)$               & 13.78 &  12.43 &  86.49 &  9.40  & 272.66 & 27.10 & 46.12 & 28.14 \\
\;-$\log \ps(\x)$             & 13.82 &  12.31  &  86.92  &  9.40   & 272.71 & 27.30 & 46.98 & 28.22  \\
\;-$2\log Z$                  & 0.20 &  0.27  &  0.56  &  0.09     &  1.97  & 1.87 & 0.41 & 0.54  \\
\;  ess             & 81.5\% &  89.1\%  &  11.2\% &  92.5\%    &  16.8\% & 22.5\% & 70.6\% & 55.9\% \\
\hline
\end{tabular}
\caption{
Negative log-likelihood (NLL) on various binary datasets from the UCI repository:
The top rows quote results from shallow models with
autoregressive weights between their units within one layer. The second block
shows results from non-autoregressive models (quoted from \citet{BornscheinBengio2015}).
In the third block we show the results obtained by training a BiHMs.
We report the estimated test set NLL when evaluating just the top-down
model, $\log p(\x)$, and when evaluating $\log \ps(\x)$. Our BiHM models
consistently obtain similar or better results than RWS while they prefer
deeper architectures. All effective sample size (see \ref{sec:ess}) estimates
are with error bounds smaller than $\pm 0.2 \%$. }
\label{tab:other}
\end{table*}

\paragraph{Sampling and inpainting}
\label{sec:inpainting}

We now discuss two general approaches for approximate sampling
from a BiHM.  One can either easily and efficiently sample from the directed
model $p$, or one can use Gibbs sampling to draw higher-quality samples from the undirected
model $\ps$. For the latter, importance resampling is used to approximately
draw samples from the conditional distributions, e.g. from:
\begin{align*}
  \ps(\h_1^{\kk}| &\x, \h_2) = \\
   &\frac{\sqrt{ p(\h_1^\kk|\h_2) p(\x|\h_1^\kk) q(\h_1^\kk|\x) q(\h_2|\h_1^\kk)}}{\sum_{\h_1} \sqrt{p(\h_1|\h_2) p(\x|\h_1) q(\h_1|\x) q(\h_2|\h_1)}}
   \enspace.
\end{align*}
Here we choose to draw the proposal samples from the mixture
distribution  $\sfrac{1}{2} \, p(\h_1|\h_2) + \sfrac{1}{2} \, q(\h_1|\x)$,
which ensures that we have a symmetric chance of covering the high
probability configurations of $\ps(\h_1|\x, \h_2)$ induced by $p$ and $q$. We
resample a final sample from $\ps(\h_1|\x, \h_2)$ propotionally to their
importance weights which are thus given by
\begin{align*}
\omega^\kk = C \frac{\sqrt{ p(\h_1^\kk|\h_2) p(\x|\h_1^\kk) q(\h_1^\kk|\x) q(\h_2|\h_1^\kk)}}
               {p(\h_1^{(k)}|\h_2) + q(\h_1^{(k)}|\x)} \enspace,
\end{align*}
%
%\begin{equation*}
%\omega_k=\frac{\sqrt{ p(\h_l^\kk|\h_{l+1}) p(\h_{l-1}|\h_l^\kk) q(\h_l^\kk|\h_{l-1}) q(\h_{l+1}|\h_l^\kk)}}{ p(\h_l^{\kk}| \h_{l+1}) + q(\h_l^{\kk}| \h_{l-1})} \cdot C \enspace,
%\end{equation*}
%where $C= \frac{2}{ \sum_{\h_l} \sqrt{p(\h_l|\h_{l+1}) p(\h_{l-1}|\h_l) q(\h_l|\x) q(\h_{l+1}|\h_l)}}$ is independent of the sample $\h_l^\kk$ and thus can be ignored when drawing samples from
%$\{\h_l^{(1)}, \h_l^{(2)}, \dots, \h_l^{(K)}\}$ with a probability proportional to $\{\omega_1,\omega_2, \dots, \omega_K\}$.
%
where $\h_1^{(k)}$ is randomly drawn from $p(\h_1|\h_2)$ or $q(\h_1|\x)$ and the
constant $C$ collects all terms not containing $\h_1^\kk$ and can be ignored.
%  when
% drawing a final sample from $\{\h_1^{(1)}, \h_1^{(2)}, \dots, \h_1^{(K)}\}$ with a
% probability proportional to $\{\omega_1,\omega_2, \dots, \omega_K\}$.
For $\ps(\x|\h_1)$ we choose to sample by drawing the proposal samples from
$p(\x|\h_1)$. We iteratively update all odd layers followed
by all even layers until we consider the chain to be in equilibrium.

% (pseudo code can be found in algorithm \ref{alg:resampling} in Appendix \ref{sec:AppendixIRwights} ).
% We then successively update all layers drawing $\h_l^{(m+1)} \sim \ps(\h_l|\h_{l-1}^{(m)}, \h_{l+1}^{(m)})$ like
% in Gibbs-sampling \af{where we can update all even}.
% We here perform multiple up- and downward sweeps until we
% consider the chain converged.
%\af{[Check: Dont we start usually with k=1?]}

Equipped with approximate sampling procedures for the conditional distributions,
it is straightforward to construct an algorithm for inpainting: Given a
corrupted input datapoint $\tilde{\x}$, we first initialize a Markov
chain by drawing $\h_1, \h_2 \sim q(\h_1, \h_2|\x)$ and then run the Gibbs sampling
procedure. Whenever we sample the bottom layer
$\x \sim \ps(\x|\h_1)$ %(approximately)
, we keep the non-corrupted elements of
$\tilde{\x}$ fixed.
%\jbrm{Note that this method approximately samples reconstructions
%$\x \sim \ps(\x)$ that are consistent with $\tilde{\x}$; it does not provide
%a MAP reconstruction which would maximize $\log \ps(\x)$ given
%$\tilde{\x}$.}
% b) Typical estimates for $2 \log Z$ range between $-4$ and $0$.
% In this range the logarithm is still a {\em relatively linear} function. The
% variance of the $Z^2$ estimate will thus only translate to moderately strong
% bias for $2 \log Z$.
%
% \iffalse
% Another way to obtain an IS estimate of $\ps$ is to evaluate the generative $p$
% model instead of $\pts$, again using Q as proposal:
% %
% \begin{align}
%   \ph(\x) &\simeq \E{ \substack{\h_2 \sim q(\h_2|h_1) \\ \h_1 \sim q(\h_1|\x)}}
%                  { \frac{p(x, \h_1, \h_2)}{q(\h_1 | \x) \; q(\h_2 | \h_1)}}
% \end{align}
% %
% Here we obtain an unbiased estimator for $pts$. When computing the LL in expectation
% over some dataset $\DD$ ($\E{\x\sim\DD}{\log \ph(x)})$ we thus get a consistent,
% conservative estimator which tends to underestimate the LL.
% Together with the (probably) biased $\p*hs(x)$ we might get some nice
% c*onvergence plots... maybe.
% \fi

\section{Experimental results}
\label{sec:experiments}

In this section we present experimental results obtained when applying the
algorithm to various binary datasets.  Our main goal is to
ensure that the theoretical properties discussed in section
\ref{sec:theconstruction} translate into a robust algorithm that yields
competitive results even when used with simple sigmoid belief network layers as
conditional distributions. We train all models using Adam~\citep{kingma2014adam}
with a mini-batch size of 100. We initialize the
weights according to \citet{GlorotAISTATS2010-small}, set the biases to
-1, and use $L_1$ regularization $\lambda$=$10^{-3}$ on all the weights.
Our implementation is available at % {\tt https//..(anonymized)..}
{\tt https://github.com/jbornschein/bihm}.

\paragraph{UCI binary datasets}

To ascertain that importance sampling based training of BiHMs works
in general, we applied it to the 8 binary datasets from the UCI dataset
repository that were evaluated e.g. in \citep{Larochelle+Murray-2011}. We
use a learning rate of $10^{-2}$ or $10^{-3}$ for all the experiments. The architectures, layer sizes and final LL estimates can be
found in tables \ref{table:uci} and \ref{table:uci-details}.

\begin{table}[h!]
\label{table:uci-details}
\centering
\scriptsize
\begin{tabular}{|l|c|c|}
\hline
  \bf Dataset     & \bf BiHM layer sizes & \bf RWS layer sizes \\ %  & \bf ess \\
\hline
\;{ADULT}             & 100, 70, 50, 25                 & 100, 20, 5    \\ %& $81.50\%$ \\ % \pm 0.11$ \\
\;{CONNECT4 }         & 300, 110, 50, 20                & 150, 50, 10   \\ %& $89.17\%$ \\ % \pm 0.05$ \\
\;{DNA}               & 200,150,130,100,70,50,30,20,10  & 150, 10     \\ %  & $11.28\%$ \\ % \pm 0.07$ \\
\;{MUSHROOM}          & 150, 100, 90, 60, 40, 20        & 150, 50, 10  \\ % & $92.56\%$ \\ % \pm 0.12$ \\
\;{NIPS-0-12}         & 200, 100, 50, 25                & 150, 50, 10  \\ % & $16.83\%$ \\ % \pm 0.2$  \\
\;{OCR}               & 600, 500, 100, 50, 30, 10       & 300, 100, 10 \\ % & $22.58\%$ \\ % \pm 0.09$ \\
\;{RCV1}              & 500, 120, 70, 30                & 200, 50, 10   \\ %& $70.60\%$ \\ % \pm 0.08$ \\
\;{WEB}               & 650, 580, 70, 30, 10            & 300, 50, 10   \\ %& $55.95\%$ \\ % \pm 0.15$ \\
\hline
\end{tabular}
\caption{
 Architectures for our best UCI BiHM models compared to our best RWS models. We observe that BiHMs prefer
 significantly deeper architectures than RWS.
}
\end{table}

\paragraph{Binarized MNIST}

We use the MNIST dataset that was binarized according to
\cite{MurraySal09} and which we downloaded in binarized form
~\citep{LarochelleBinarizedMNIST}. Compared to RWS, we again observe that
BiHMs prefer significantly deeper and narrower models. Our
best model consists of 1 visible and 12 latent layers with 300,200,100,75,50,35,30,25,20,15,10,10
latent variables. We follow the same experimental procedure as in the RWS paper:
First train the model with $K$=10 samples and a learning rate of $10^{-3}$ until
convergence and then fine-tune the parameters with $K$=100 samples and a learning
rate of $3 \times 10^{-4}$. All layers are actually used to model the
empirical distribution; we confirmed that training shallower models (obtained by leaving out individual layers)
decreases the performance.
We obtain test set log-loglikelihoods of $\log \ps(\x) \simeq$ {\bf -84.6 $\pm$ 0.23} and
$\log p(\x) \simeq$ {\bf -84.3 $\pm$ 0.22}
\footnote{Recently, a version of MNIST where binary observed variables are resampled during training
has been used (e.g., \citet{burda2015importance}). On this dataset we obtain $\log p(\x) \simeq$ {\bf -82.9}}.
The next section presents a more detailed analysis
of these estimates and their dependency on the number of samples from the proposal distribution $q(\h|\x)$. Note that even though this model is relatively deep, it is  not particularly large,
with about $700,000$ parameters in total. The DBMs in
\cite{Salakhutdinov2009deep} contain about $900,000$ and 1.1 million
parameters; a variational autoencoder with two deterministic, 500 units wide
encoder and decoder layers, and with 100 top level latent units contains more than
1.4 million parameters.

To highlight the models ability to generate crisp (non-blurry) digits we draw approximate samples
from $\ps(x)$ which are visualized in Fig. \ref{fig:mnist} {\bf B}. Fig. \ref{fig:mnist} {\bf A}
shows samples obtained when drawing from the top-down generative model
$p(\x)$ before running any Gibbs iterations. Fig. \ref{fig:inpainting}
visualizes the results when running the inpainting algorihm to reconstruct
partially occluded images. Our sourcecode package contains additional results and
animations.

\begin{table}[h]
\label{table:mnist}
\centering
\scriptsize
\begin{tabular}{|l|c|c|}
\hline
  \bf Model &    $\le$  - $\log p(x)$ &    $\approx$ - $\log p(\x)$ \\
\hline
{\bf autoregressive models} &      &      \\
\; NADE                           &   $-$   & 88.9 \\
\; DARN (1 latent layer)          & 88.3 & 84.1 \\
{\bf continuous latent variables} &      &      \\
\; VAE                            & 96.2 & 88.7 \\
\; VAE + HMC (8 iterations)       & 88.3 & 85.5 \\
\; IWAE (2 latent layers)         & 96.1 & 85.3 \\
%\hline
{\bf binary latent variables}     &      &      \\
\; NVIL (2 latent layers)         & 99.6 &  $-$    \\
\; RWS  (5 latent layers)         & 96.3 & 85.4 \\
\; {\bf BiHM} (12 latent layers)  & {\bf 89.2} & {\bf 84.3} \\
\hline
\end{tabular}
\caption{Comparison of BiHMs to other recent methods in the literature. We report
 the lower bounds and estimates for the marginal log probability on the binarized MNIST
 test set.}
\end{table}

\begin{figure*}[t]
 \includegraphics[width=1.0\linewidth]{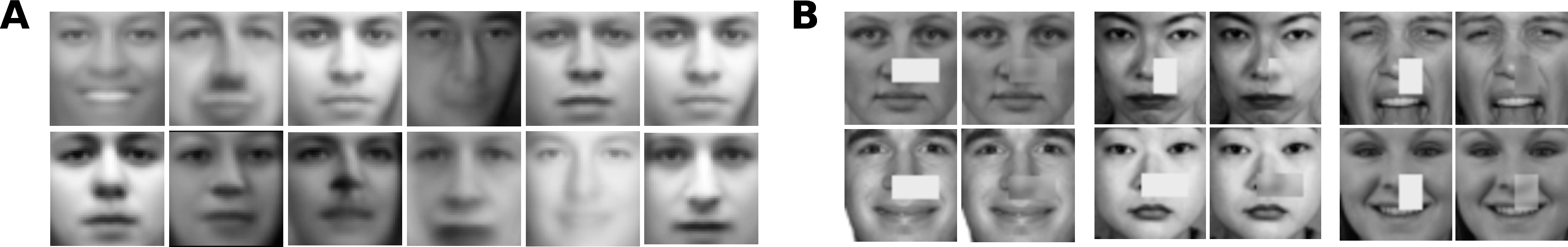}
 \vspace{-0.7cm}
 \caption{Results after training on TFD: {\bf A}) Random selection of 12 samples
 drawn from $\ps(\x)$ (10 iterations of Gibbs sampling).
 {\bf B}) The left column in each block shows the input; the right column shows
 a random output sample generated by the inpaiting algorithm (see section \ref{sec:inpainting}).
 }
 \label{fig:tfd}
\end{figure*}

\paragraph{Toronto Face Database}
We also trained models on the 98,058 examples from the unlabeled section of
the Toronto face database~(TFD, \citet{Susskind2010}). Each training example is of size
$48\times48$ pixels and we interpret the gray-level as Bernoulli
probability for the bottom layer. We observe that training proceeds
rapidly during the first few epochs but mostly only learns the
mean-face. During the next few hundred epochs training proceeds much slower but
the estimated log-likelihood $\log \ps(\x)$ increases steadily.
Fig. \ref{fig:tfd} {\bf A} shows random samples from a model with respectively
1000,700,700,300 latent variables in the 4 hidden layers. It was trained with
a learning rate of $3 \cdot 10^{-5}$; all other hyperparameters were set to the
same values as before. Fig. \ref{fig:tfd} {\bf B} shows the results from
inpainting experiments with this model.

\begin{figure*}[t]
 \includegraphics[width=1.0\linewidth]{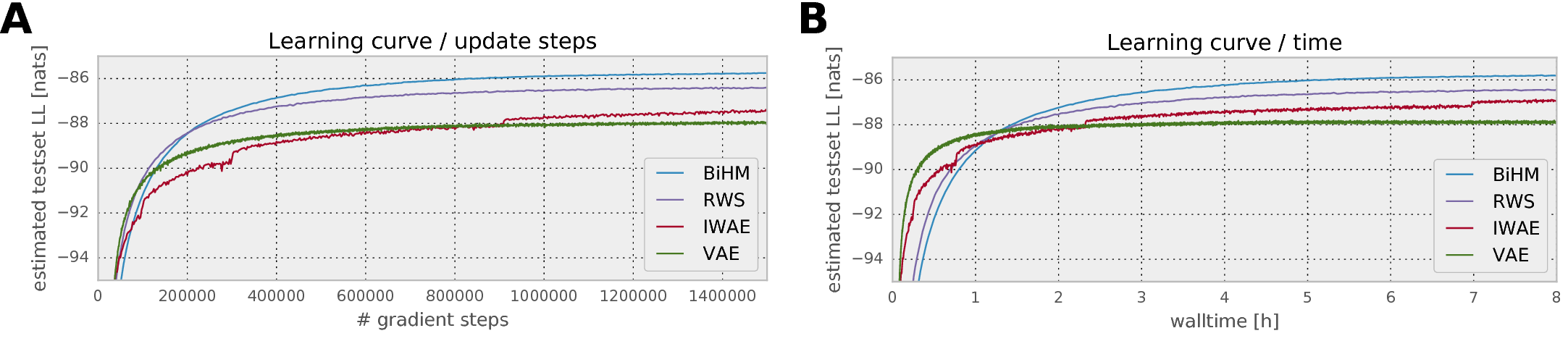}
 \vspace{-0.7cm}
 \caption{
 Learning curves for MNIST experiments: For BiHM, RWS and VAE we
 chose the learning rate such that we get optimal results after $10^6$ update
 steps; for IWAE we use the original learning rate schedule published in
 \citep{burda2015importance}. BiHM and RWS use $K$=10 samples per datapoint;
 IWAE uses $K$=5 and a batch size of 20 (according to the original publication).
 We generally observe that BiHM show very good convergence in terms of progress
 per update step and in terms of total training time.
 }
\label{fig:lcurves}
\end{figure*}

\subsection{Analysis of importance sampling-based estimates}

\paragraph{Estimating the partition function}
In Fig. \ref{fig:zest} {\bf A} we plot $2 \log Z$ estimates (equation
\eqref{eqn:z-est}) over the number of outer samples $K_{\text{outer}}$ for our
best MNIST model and for 3 different choices of $K_{\text{inner}}$, i.e.,$K_{\text{inner}} \in \{1, 10,
100\}$. % The gray area additionally shows the  mean and standard deviation for 10 runs with $K_{\text{inner}}=1$.
In Fig. \ref{fig:zest} {\bf B} we plot the estimates over the total number of
samples $K_\text{outer} \cdot K_\text{inner}$. We observe that choosing
$K_\text{inner}$ =1 and using only about 10 million samples results in high
quality estimates for $2 \log Z$ with an standard error far below 0.1  nats.
Estimating based on 10 million samples takes less than 2 minutes on a GTX980
GPU. Fig. \ref{fig:zest} {\bf C} shows the development of the $2 \log Z$
estimate during learning and in relation to the LL estimates.

\paragraph{Importance sampling efficiency}
\label{sec:ess}
A widely used metric to estimate the quality of an importance sampling estimator
is the {\em effective sampling size} (ESS), given by
$\widehat{ess} = \sfrac{(\sum_{k=1}^K \omega_k)^2}{(\sum_{k=1}^K \omega_k^2)}$
(see, e.g., \citet{RobertCasella2009}). Larger values indicate
more efficient sampling (more information extracted per sample).
%
% \begin{align}
%  \widehat{ess} = \frac{(\sum_{k=1}^K \omega_k)^2}{\sum_{k=1}^K \omega_k^2}
%  \enspace.
% \end{align}
%
We compute the ESS over the MNIST test set for
$K$=100,000 proposal samples from $q(\h|\x)$. For our best RWS model, a model
with 5 stochastic layers (400,300,200,100,10), we obtain $\widehat{ess} \simeq
0.10\% \pm 0.06$; for the BiHM model we obtain $\widehat{ess} \simeq 11.9\% \pm
1.1$. When we estimate the ESS for using $q(\h|\x)$ from the BiHM as a proposal
distribution for $p(\h|\x)$, we obtain $\widehat{ess}$=$1.2\% \pm 0.2 $. The estimated ESS values indicate that
training BiHM models indeed results in distributions whose intractable posterior
$\ps(\h|\x)$ as well as top-down model $p(\h|\x)$ are much better modeled by the learned $q(\h|\x)$.
%\jbrm{Although not directly comparable, }
We also estimated the ESS for a VAE with two determninistic, 500 units wide ReLU layers in the encoder
and decoder. This model has a single stochastic layer with 100 continuous
variables at the top; it reaches a final estimated test set LL of $\log p(\x)
\simeq -88.9 \pm 0.28$.
The final variational lower bound, which corresponds exactly to the
importance sampling estimate of $\log p(\x)$ with $K$=1 sample, is $-95.8$. For this model we obtain
an ESS of $0.07\% \pm 0.02$. These results indicate that we need thousands of samples to obtain
reliable LL estimates with low approximation error.
%\af{[Check: Soll sich der Satz auf alle Modelle oder auf VAEs beziehen? Wir sagen ja nachher, dass bei BiHM auch weniger genuegen?]}
In Fig. \ref{fig:mnist} {\bf C} we plot the estimated test set LL over
the number of samples $K$ used to estimate $\log \ps(\x)$ and $\log p(\x)$.
For all the models and for small a number of samples $K$ we significantly
underestimate the LL; but, in comparison to RWS, the estimates for the BiHM
model are much higher and less sensitive to $K$.
E.g, using $K$=10 samples to evaluate the BiHM model results in a higher LL
estimate than using $K$=10,000 samples to evaluate the RWS model.

\paragraph{Computational cost}
\label{sec:comp}
To demonstrate the computational efficiency of our approach we show typical
MNIST learning curves in  Fig. \ref{fig:lcurves}. For BiHM, RWS and VAE the
learning rate was chosen within a factor of 2 to obtain optimal results after
$10^6$ update steps ($5 \cdot 10^{-4}$ for BiHM and RWS, $3 \cdot 10^{-3}$ for VAE; $K$=10 for BiHM and RWS).
For the IWAE experiment we use the original code, hyperparameters and learning rate
schedule from \citep{burda2015importance}: This experiment thus uses a mini-batch size
of 20 instead of 100, $K$=5 training samples and and 8 different learning rates
over the course of $\approx$ 3300 epochs. In all cases we used $K$=1000 samples
to evaluate the test set log-likelihoods.
We generally observe that BiHM show very good convergence in terms of
progress per update step and competitive performance in terms of total training time.
Note that BiHMs and RWS allow for an efficient distributed implementation in
the future: per sample, only the binary activations and a single floating
point number (the importance weight) need to be communicated between layers.
VAEs and IWAEs need to communicate continuous activations during the forward
pass and continuous partial gradients during the backward pass.
At test time BiHMs are typically much more effective than
the other methods: BiHMs obtain good LL estimates with
$K$=10 or 100 samples per datapoint while VAE, RWS and IWAE models
require $\approx$ 10,000 samples to obtain competitive results
(compare Fig. \ref{fig:mnist} {\bf C}).

\section{Conclusion and future work}
\label{sec:conclusion}

We introduced a new scheme to construct probabilistic generative models which
are automatically regularized to be close to approximate inference
distributions we have at our disposal. Using the Bhattacharyya distance we
derived a lower-bound on the log-likelihood, and we demonstrated that the bound
can be used to fit deep generative models with many layers of latent variables
to complex training distributions.

Compared to RWS, BiHM models typically prefer many more latent layers. After training
a BiHM, the directed top-down model $p$ shows better performance than
a RWS trained model; both in terms of log-likelihood and sample quality.
Sample quality can be further improved by approximately sampling from the
full undirected BiHM model $\ps$. The high similarity between $\ps$ and $q$,
enforced by the training objective, allows BiHMs to be evaluated
orders of magnitude more efficiently than RWS, VAE and IWAE models.

% Note that our definition for $\ps$ forced
% us to choose a prior distribution $q(\x)$ which will be part of our generative
% model $\ps(\x, \h)$. This is different from the typical variational approaches
% to train Helmholtz machines where we would think of $q(\h|\x)$ solely as an
% approximate inference method given a training example $\x$, and where $q(\x)$
% would be the (empirical) training distribution -- something we cannot assume
% because $q(\x)$ is part of our model $\ps$.

Possible directions for future research could involve semisupervised learning:
the symmetric nature of the generative model $\ps$ (it is always close to the
bottom-up and top-down directed models $q$ and $p$) might make it particularly interesting
for learning tasks that require inference given changing sets of observed
and hidden variables.
We also have a wide range of potential choices for our
parametrized conditional distributions. Assuming continuous latent variables
for example and eventually choosing an alternative inference method might make
$\ps$ a better suited model for some training distributions.

{\bf Acknowledgments}: \\
We thank the developers of Theano~\cite{bergstra+al:2010-scipy-small} and
Blocks~\cite{Van+al-arxiv-2015} for their great work. We thank NSERC, Compute
Canada, CIFAR and Samsung for their support.

%\subsubsection*{References}
%\newpage
\bibliography{strings,strings-shorter,ml,aigaion,localref}
\bibliographystyle{iclr2016_conference}

\end{document}